# Automatic Annotation of Multilingual Text Collections with a Conceptual Thesaurus

**Bruno Pouliquen, Ralf Steinberger, Camelia Ignat**
European Commission - Joint Research Centre (JRC)
Institute for the Protection and Security of the Citizen (IPSC)
T.P. 267, 21020 Ispra (VA), Italy
http://www.jrc.it/langtech
Firstname.Lastname@jrc.it

## Abstract

Automatic annotation of documents with controlled vocabulary terms (*descriptors*) from a conceptual thesaurus is not only useful for document indexing and retrieval. The mapping of texts onto the same thesaurus furthermore allows to establish links between similar documents. This is also a substantial requirement of the Semantic Web. This paper presents an almost language-independent system that maps documents written in different languages onto the same multilingual conceptual thesaurus, EUROVOC. Conceptual thesauri differ from Natural Language Thesauri in that they consist of relatively small controlled lists of words or phrases with a rather abstract meaning. To automatically identify which thesaurus descriptors describe the contents of a document best, we developed a statistical, associative system that is trained on texts that have previously been indexed manually. In addition to describing the large number of empirically optimised parameters of the fully functional application, we present the performance of the software according to a human evaluation by professional indexers.

## 1 Introduction

The process of assigning keywords to documents is called *indexing*. It is different from the process of producing an inverted index of all words occurring in a text, which is called *full text indexing*. Lancaster (1998) distinguishes the indexing tasks *keyword extraction* and *keyword assignment*. *Keyword extraction* is the task of identifying keywords present verbatim in text, while *keyword assignment* is the identification of appropriate keywords from the controlled vocabulary of a reference list (a *thesaurus*). Controlled vocabulary keywords, which are usually referred to as *descriptors*, are therefore not necessarily present explicitly in the text.

We furthermore distinguish *conceptual thesauri* (CT) from *natural language thesauri* (NLT). In CT, most descriptors are relatively abstract, conceptual terms. An example for a CT is EUROVOC (Eurovoc, 1995; see section 1.2), whose approximately 6,000 descriptors describe the main concepts of a wide variety of subject fields by using high-level descriptor terms such as PROTECTION OF MINORITIES, FISHERY MANAGEMENT and CONSTRUCTION AND TOWN PLANNING [1]. NLT, on the other hand, are more concrete in the sense that they usually aim at including an *exhaustive* list of the terminology of the covered field. Examples are MeSH in the medical field (NLM, 1986), DESY in particle physics (DESY, 1996), and AGROVOC in agriculture (AGROVOC, 1998). WordNet (Miller, 1995) is a NLT that is not specialised in any particular subject domain, but it does have the aim of being exhaustive (distinguishing approximately 95,000 synonym sets).

In this paper, we present work on automating the process of keyword *assignment* in several languages, using the CT EUROVOC. The challenge of this task is that the EUROVOC descriptor texts are not usually explicitly present in the documents. We can therefore show in section 3 that treating EUROVOC descriptor identification as a keyword *extraction* task leads to very bad results.

---

[1]We write all EUROVOC descriptors in small caps.

We succeeded in making the big step from keyword extraction to keyword assignment by devising a statistical system that uses a training corpus of manually indexed documents to produce, for each descriptor, a list of associated natural language words whose presence in a text indicates that the descriptor may be appropriate for this text.

## 1.1 Contents

The structure of this paper is the following: we first present the EUROVOC thesaurus and explain why so many organisations use thesauri instead of, or in addition to, using conventional full-text search engines. In section 2, we then distinguish our system from related work. In section 3, we give a high-level overview of the approach we adopted, without specifying the details. The reason for keeping the description general is that we experimented with many different formulae, parameters and parameter settings, and these will be listed in sections 4 to 6. Section 4 discusses the experiments concerning the *linguistic preprocessing* of the texts and the various results achieved. Section 5 is dedicated to those parameters that were used during the *training phase* of the process to produce the most efficient list of associated words for each descriptor. Section 6 then discusses the various experiments carried out to optimise the *descriptor assignment* results by matching the associated words against the text, to which descriptors should be assigned.

Section 7 summarises the results achieved with the best parameter settings according to a manual evaluation by indexing professionals. The conclusion summarises the findings, shows possible uses of our system for other applications, and points to future work.

## 1.2 The Eurovoc thesaurus

EUROVOC (Eurovoc, 1995) is a wide-coverage conceptual thesaurus, covering diverse fields such as politics, law, finance, social questions, science, transport, environment, geography, organisations, etc. EUROVOC is used by the European Parliament, the European Commission's Publications Office and at least fifteen other (mostly parliamentary) institutions to catalogue their multilingual document collections for search and retrieval. It exists in one-to-one translations in eleven languages with a further eleven language versions awaiting release. EUROVOC descriptors are defined precisely, using *scope notes*, so that each descriptor has exactly one translation into each language. A dedicated maintenance committee continuously updates the thesaurus.

EUROVOC is a thriving resource that will be used by more organisations in the future as it facilitates information and document exchange between parliamentary and other databases in the European Union, its Member States and other countries. It is also likely that more EUROVOC language versions will be developed.

EUROVOC organises its 6075 descriptors hierarchically into eight levels, using the relations *Broader Term* and *Narrower Term* (BT/NT), as well as *Related Term* (RT). RTs link nodes not related hierarchically. EUROVOC also provides a number of language-specific and optional *non-descriptor* terms that may help the indexing professional to find the appropriate descriptor. Non-descriptors typically are synonyms or hyponyms of the descriptor term (e.g. *banana* for TROPICAL FRUIT).

## 1.3 Motivation for thesaurus indexing

Most large organisations use thesauri for consistent indexing, storage and retrieval of electronic and hardcopy documents in their libraries and documentation centres. A list of carefully chosen descriptors gives users a quick summary of the document contents and it enables them to navigate the document collection by subject field. The hierarchical nature of the thesaurus allows the query expansion in database retrieval of documents by subject field (e.g. 'radioactive materials') without having to enter a list of possible search terms (e.g. 'plutonium', 'uranium', etc.). When using multilingual thesauri such as EUROVOC, multilingual document collections can be searched monolingually by taking advantage of the fact that there are one-to-one translations of each descriptor.

Manual assignment of thesaurus descriptors is time-consuming and expensive. Several organisations confirmed that their professional EUROVOC indexers assign, on average, less than thirty documents per day. Thus, automatic or, at least semi-automatic, solutions are sought. The JRC system takes about five seconds per document and could be used as a fully-automatic system or as input to machine-aided indexing.

Apart from supporting organisations that currently use manually assigned EUROVOC descriptors, the automatic descriptor assignment can be useful to catalogue other types of documents, and for several other purposes: Representing document contents by a list of multilingual descriptors allows multilingual document classification and clustering, cross-lingual document similarity calculation (Steinberger et al., 2002), and the production of multilingual document maps (Steinberger, 2000). Lin and Hovy (2000) showed that the data produced in a similar process can also be useful for subject-specific summarisation. Last but not least, linking texts to meta-information such as established thesauri is a prerequisite for the realisation of the *Semantic Web*. EUROVOC is getting more widely accepted as a standard for parliamentary documentation centres. Due to its wide coverage, its usage is in no way restricted to parliamentary texts. As it will also soon be available in 22 languages, EUROVOC has the potential for being a good standard reference to link documents on the Semantic Web.

## 2 Related work

Most previous work in the field concerns the indexing of texts with specialised natural language thesauri. These efforts come closer to the task of keyword *extraction* than keyword assignment because exhaustive terminology lists exist that can be matched against the words in the document to be indexed. Examples are Pouliquen et al. (2002) for the field of medicine, Montejo-Raez (2002) for particle physics and Haller et al. (2001) for economics. Jacquemin et al. (2002) additionally used tools to identify morphological and syntactic variations of the descriptors of the agricultural thesaurus AGROVOC. Gonzalo et al.'s (1998) effort to identify the most appropriate WordNet synsets for a text also differs from our own work: While the major challenge for WordNet indexing is to sense-disambiguate words found in the text that are part of several synsets, EUROVOC indexing is difficult because the descriptors are *not* present in the text.

Regarding indexing with *conceptual* thesauri (CT), both Marjorie & Hlava (1996) and Loukachevitch & Dobrov (2002) use rule-based approaches using vast, language-specific linguistic resources. Marjorie & Hlava's system to assign English EUROVOC descriptors uses over 40,000 hand-crafted rules making use of text strings, synonym lists, vicinity operators and even tools to recognise and exploit legal references in text. Such an excessive usage of language-specific resources is out of our reach as we aim at linguistics-poor methods so that we can adapt them to all Eurovoc languages.

The most similar application to ours was developed by Ferber (1997), whose aim was to use a multilingual thesaurus for the retrieval of English documents using search terms in languages other than English. Ferber trained his associative system on the titles of 80,000 bibliographic records, which were manually indexed using the OECD thesaurus. The OECD thesaurus is similar to EUROVOC, with the difference that it is smaller and exists only in four languages. Ferber achieved rather good results (a precision of 62% for a recall of 64%). However, we cannot compare our methods and our results directly with his as the training data is of a rather different nature (corpus of titles vs. corpus of full texts with highly varying length).

Our approach of producing lists of associated words whose presence in a text indicate the appropriateness of the corresponding descriptor is not dissimilar to work on *topic signatures*, as described in Lin and Hovy (2000) and Agirre et al. (2000). However, our application requires a few additional steps because it is more complex and there are also a number of differences regarding the creation of the lists. Lin and Hovy produced their topic signatures on documents that had been classified manually as being or not being relevant for one of four specific domains. Also, Lin and Hovy used the topic signatures to relevance-rank sentences in text of the same domain for the purpose of summarisation. They were thus able to use positive and negative training examples and they only had to decide, to what extent a sentence is similar to their topic signature, separately for each of the four subject domains. Agirre et al. did produce topic signatures for many more subject domains (for all WordNet synsets), but they used the signatures for word sense disambiguation, meaning that for each word they had only as many choices as there were word senses.

In the case of EUROVOC descriptor assignment, the situation is rather different, in that, for each document, it is possible to assign any of the 6075 descriptors and, in fact, multiple classification is the aim. Additionally, descriptor lists are required to be short so that only the most relevant descriptors should be assigned while other appropriate, but less relevant descriptors should not be assigned in order to keep the list concise.

Due to the complexity of this task, we introduced a large number of additional parameters that were not used by the authors mentioned above. Some of these parameters concern the pre-processing of the texts, some of them affect the creation of the topic signatures, and again others were introduced to optimise the mapping of the signatures with the texts for which EUROVOC descriptors are sought. Sections 4 to 6 explain these parameters in detail.

## 3 Overview of the process

### 3.1 Test and training corpus

Our English corpus consists of almost 60,000 texts of eight different types[2]. Documentation specialists had indexed them manually with an average of 5.65 descriptors per text over a period of nine years. For some EU languages, the training corpus is slightly smaller. The average text size is about 5,500 characters, with a rather high standard deviation of 17,000. We randomly selected 587 texts to build a test set that is representative of this corpus regarding the various text types. The remainder was used for training.

### 3.2 'Extracting' EUROVOC descriptors

The analysis of the training corpus showed that only 31% of the documents contain explicitly the manually assigned descriptor terms. At the same time, in nine out of ten cases where a descriptor text occurred explicitly in a text, this descriptor was *not* assigned manually. These facts indicate that identifying the most appropriate EUROVOC descriptors by keyword *extraction* (i.e. solely by searching for their verbatim occurrence in the text) will not yield good results.

To prove this, we launched an extraction experiment on our English test set. We assigned all descriptors automatically whose descriptor text occurred explicitly in the document. In order to evaluate the outcome, we compared the results with those EUROVOC descriptors, that had previously been assigned manually to these texts. The experiment showed that a maximum *Recall* of 30.8% could be achieved, i.e. almost 70% of the manually assigned descriptors were not found. At the same time, this method achieved a precision of 7.4%, meaning that over 92% of the automatically assigned descriptors had not been assigned manually. We also experimented with using a lemmatiser, stop words (as described in section 4) and EUROVOC's non-descriptors. These experiments never yielded better Precision values, but the maximum Recall could be elevated to 39.8%.

These are very poor results, which prove that keyword *extraction* is indeed not an option for the EUROVOC thesaurus. We take this performance as a lower-bound benchmark, assuming that our system has to perform better than this.

### 3.3 'Assigning' EUROVOC descriptors, using an associative approach

As keyword extraction is not an option, we adopted a linguistics-poor statistical approach and trained a system on our corpus. As the only types of linguistic input, we experimented with normalising all texts of the training and test sets, using lemmatisation, multi-word mark-up and removing stop words (see section 4).

During the training phase, we produce a ranked list of words (or: *lemmas*) that are statistically (and often also semantically) related to each descriptor (see section 5). We refer to these lemmas as *associates*. These associate lists are rather similar to the *topic signatures* mentioned in section 2. **Table 1** shows an example associate list for the EUROVOC descriptor FISHERY MANAGEMENT. The various columns will be explained in section 5.

During the assignment phase, we normalise the new document in the same way and calculate the similarity between this document's lemma frequency list and each of the descriptor associate lists (see section 6). The descriptor associate lists that are most similar to the lemma frequency

---

[2] Types of document are 'Parliamentary Question', 'Council Regulation', 'Council Decision', 'Resolution, 'Protocol', 'Debate', 'Contract', etc.

| Lemma | Freq | Nb of texts | Weight |
|-------|------|-------------|--------|
| fishery_resource | 317 | 160 | 54.47 |
| fishing | 983 | 281 | 49.11 |
| fish | 1766 | 281 | 46.19 |
| common_fishery_policy | 274 | 165 | 44.67 |
| fishery | 1427 | 281 | 44.19 |
| fishing_activity | 295 | 124 | 43.37 |
| fly_the_flag | 403 | 143 | 42.87 |
| aquaculture | 242 | 171 | 39.27 |
| conservation | 759 | 183 | 38.34 |
| vessel | 2598 | 230 | 37.91 |
| ... | | | |

**Table 1**. Top ten associated lemmas for EUROVOC descriptor FISHERY MANAGEMENT. With reference to the discussion in section 5, the columns 2, 3 and 4 show the absolute *frequency* of the lemma in all texts indexed with this descriptor, the number of *texts* indexed with this descriptor the lemma occurred in, and the final *weight* of each lemma.

list of the new document indicate the most appropriate EUROVOC descriptors.

The EUROVOC descriptors can then be presented in a ranked list, according to the similarity of their associate lists with the documents' lemma frequency list, as shown in **Table 2**. As the list of potential descriptors is very long, we must decide how many descriptors to present to the users, and for how many descriptors to calculate Precision and Recall values. We can compute Precision and Recall for any number of highest-ranking descriptors. If we say that the Precision at rank 5 is Y%, this means that an av-erage of Y% of the top five descriptors were correct in all documents evaluated. During the training phase, we evaluated the automatically generated descriptor lists automatically, by comparing them to the previously manually assigned descriptors. The more manually assigned descriptors were found at the top of the ranked list, the better the results. The final evaluation of the assignment, as discussed in section 7, was carried out manually.

| Rank | Descriptor | Similarity |
|------|-----------|-----------|
| 1 | VETERINARY LEGISLATION | 42.4% |
| 2 | PUBLIC HEALTH | 37.1% |
| 3 | VETERINARY INSPECTION | 36.6% |
| 4 | FOOD CONTROL | 35.6% |
| 5 | FOOD INSPECTION | 34.8% |
| 6 | AUSTRIA | 29.5% |
| 7 | VETERINARY PRODUCT | 28.9% |
| 8 | COMMUNITY CONTROL | 28.4% |

**Table 2**. Assignment results (8 top-ranking descriptors) for the document *Food and veterinary Office mission to Austria*, found on the internet at http://europa.eu.int/comm/food/fs/inspections/vi/reports/austria/vi_rep_oste_1074-1999_en.html.

For each formula and parameter, we identified the optimal parameter setting in an empirical way, by trying a range of parameters and by then choosing the setting that yielded the best results. For parameter tuning and evaluation, we carried out over 1500 tests.

We will now focus on the description of the different parameters used in the pre-processing, training and assignment phases. Section 7 will then show the results according to a human evaluation of the descriptor assignment, using an optimised parameter setting.

## 4 Corpus pre-processing

We tried to keep the linguistic effort minimal in order to be able to apply the same algorithm to all eleven languages for which we have training material. Our initial assumption was that lemmatisation would be crucial (especially for languages that are more highly inflected than English), that marking up multi-word expressions would be useful as it helps disambiguating polysemous words such as 'plant' (*power plant* vs. *green plant*), and that stop words would help excluding words that are semantically poor or that can be considered as corpus-specific 'noise'. **Table 3** shows that, for both English and Spanish, using the combination of lemmatisation, multi-word mark-up and stop word lists does indeed produce the best results. However, only using the corpus-tuned stop word list containing 1533 words yields results that are almost as good (Spanish F = 47.4 vs. 48). This result was a big surprise for us. Tuning the stop word list to the domain clearly was useful as the results achieved with a standard stop word list were less good (F = 46.6 vs. 47.4).

We conclude that, at least if the amount of training material is similar, and for languages that are not more highly inflected than Spanish, the assignment results do not suffer much if no lemmatisation and multi-word treatment is carried out. This is good news as this makes it easier to apply the algorithm to more languages, for which less linguistic resources may be available.

| LEM | SW | MW | Prec Spanish | Recall Spanish | F-measure Spanish | F-measure Engl. |
|---|---|---|---|---|---|---|
| − | − | − | 40.3 | 43.4 | 41.8 | 45.6 |
| − | − | + | 40.4 | 43.6 | 41.9 | |
| − | + | − | 45.6 | 49.3 | **47.4** | **49.1** |
| − | *strict* | − | 44.8 | 48.5 | **46.6** | |
| − | + | + | 45.6 | 49.2 | 47.3 | |
| + | − | − | 42.4 | 45.6 | 43.8 | |
| + | + | − | 45.7 | 49.6 | 47.6 | 48.5 |
| + | − | + | 43.9 | 47.4 | 45.6 | |
| + | + | + | 46.2 | 49.9 | **48.0** | **50.0** |

**Table 3**. Evaluation of the assignment (for the 6 top-ranking descriptors) on Spanish and English texts following various linguistic pre-processing steps. **LEM**: using lemmatisation, **SW**: using stop word list, **MW**: marking up multi-word expressions. "*strict*" indicates that we used a general, non-application-specific stop word list; missing results have not been computed.

## 5    Producing associate lists

The process of creating associate lists (or 'topic signatures') for each descriptor is the part where we experimented with most parameters. We can only mention the major parameters here, as explaining all details would require more space. The result of this process is, for each descriptor, a vector consisting of all associates and their weight, as shown in **Table 1**.

We did not have enough training material for all descriptors as some descriptors were never used and others were used very rarely. We distinguish (I) basic *minimum requirements* that had to be met for us to produce associate lists, or basic decisions we took, and (II) parameters that had an impact on the *choice* of associates and their weights. Using the optimised minimum requirements, we managed to produce associate lists for 2893 English and 2912 Spanish descriptors.

(I) Basic requirements and decisions:
(a)   minimum size and number of training texts available for each descriptor. We chose to

require at least 5 texts with at least 2000 characters each (half a page).

(b)   produce associate lists on the basis of one large meta-text per descriptor (concatenation of all texts indexed with this descriptor) vs. producing associate candidates for each text indexed with this descriptor and joining the results. As the training texts were of extremely varying length, the latter method produced much better results.

(c)   choice of measure to identify associates in texts, such as pure frequency (TF), frequency normalised by average frequency in the training corpus, TF.IDF, chi-square, log-likelihood. Following Kilgarriff's (1996) study, we used log-likelihood. We set the p-value as high as 0.15 so as to produce long associate lists.

(d)   choice of reference corpus for the log-likelihood formula. We chose our training corpus as a reference corpus over using an independent corpus (like the *British National Corpus* or others).

(II) Parameters with an impact on the choice and weight of associates:

(e)   the minimum number of texts per descriptor for which the lemma is an associate. We were surprised to learn that results were best when requiring the lemma to occur in a minimum of only two texts. Setting this threshold higher means getting more descriptor-specific, but also shorter associate lists.

(f)   Deciding on the weight of an associate for a descriptor. Candidates were the frequency of the lemma in all texts indexed with this descriptor, the number of texts the lemma occurs in, the sum of the log-likelihood values, etc. Best results were achieved using the number of texts indexed with this descriptor, ignoring the absolute frequency and the log-likelihood values; the log-likelihood formula was thus only used to identify associate candidates.

(g)   normalisation of the associate weight. We used a variation of the IDF formula (F3 below), by dividing by the number of descriptors for which the lemma is an associate. This punishes the impact of lemmas that are associates to many descriptors. This proved to be so important that we punished common lemmas strongly. Using a 'β' of 10 in formula F3 cancelled the impact of all associates occurring at least 10% as often as the most common associate lemma.

(h) further normalisation of the associate weight. We considered either the length of each training text or the number of descriptors manually assigned to this text. Normalisation by text length yielded bad results, but normalisation by the number of other descriptors was important to reduce the interference of other descriptors that were assigned to the same training text (see formula F2).

(i) a minimum weight threshold for each associate. Experiments showed that results are better when not considering associates with a lower weight.

(j) a minimum requirement on the number of lemmas in the associate list of a descriptor for us to assign this descriptor. Setting this parameter high increases precision as a lot of lexical evidence is needed to assign a descriptor, but it lowers recall as the number of descriptors we can assign is low. A minimum of ten associates per descriptor produced best F-measure results.

The final formula to establish the $weight_{l,d}$ of lemma $l$ as an associate of a descriptor $d$ is:

$$Weight_{l,d} = W_{l,d} \cdot IDF_l \qquad (F1)$$

*with*: $l$ being a lemma, $d$ a descriptor, $W_{l,d}$ the weight of a lemma in a descriptor, $IDF_l$ the "Inverse Descriptor Frequency".

$$W_{l,d} = \sum_{t \in T_{l,d}} \frac{1}{Nd_t} \qquad \text{see (h)} \ (F2)$$

*with*: $t$ being a text; $Nd_t$ being the number of manually assigned descriptors for text $t$; $T_{l,d}$ being the texts that are indexed by descriptor $d$ and containing lemma $l$.

$$IDF_l = \log\left(\frac{Max_{DF_l}}{\beta \cdot DF_l} + 1\right) \qquad \text{see (g)} \ (F3)$$

*with*: $DF_l$ being the descriptor frequency, i.e. the number of descriptors the lemma appears in as an associate. $Max_{DF}$ is the maximum value of $DF_l$ for all lemmas. The parameter $\beta$ is set to 10 in order to punish lemmas that occur in more than 10% of the $Max_{DF}$ value.

The whole formula is F4:

$$Weight_{l,d} = \left(\sum_{t \in T_{l,d}} \frac{1}{Nd_t}\right) \cdot \log\left(\frac{Max_{DF_l}}{\beta \cdot DF_l} + 1\right) (F4)$$

## 6 Assigning descriptors to a text

Once associate vectors such as that in Table 1 exist for all descriptors satisfying the basic requirements laid out in section 5, descriptors can be assigned to new texts by calculating the similarity between the text and the associate vectors. To this end, the text is pre-processed in the same way as the training material and lemma frequency lists are produced for the new text. An experiment working with log-likelihood values for the lemmas of the new text instead of pure lemma frequencies gave bad results.

(III) We experimented with the following *filters* before calculating the similarity:

(k) we set a threshold for the minimum number of descriptor associates that had to be present in the text to avoid that only a couple of associates with a high weight would trigger wrong descriptors. This proved to be very important. The optimal minimal occurrence is four. Using a smoothing technique by adapting this parameter flexibly to either the text length or to the number of associates in the descriptor did not yield good results.

(l) We checked whether the occurrence of the descriptor text in the new document should be required (doing this produced bad results; see section 3.2).

(IV) We tried the following similarity measures to compare the text vector with the descriptor vectors:

(m) the Cosine formula (Salton, 1989);

(n) the Okapi formula (Robertson et al., 1994);

(o) the scalar product of vectors (cosine without normalisation);

(p) a linear combination of the three formulae, as recommended by Wilkinson (1994). In all cases, this combination produced the best results, but the optimal proportions varied depending on the languages and on the other parameter settings; an average good mix of weights turned out to be 40%-20%-40% for the formulae mentioned in (m)-(n)-(o).

# 7 Manual evaluation of the assignment

In addition to comparing our automatic assignment results to previously manually assigned descriptors, we asked two indexing specialists to evaluate the automatically generated results. The purpose of this second evaluation was (a) to get a second opinion because indexers differ in their judgements on the appropriateness of descriptors, (b) to get feedback on the relevance of those descriptors that were assigned automatically, but not manually, and (c) to produce an upper bound benchmark for the performance of our system by setting the assignment overlap between human indexers as the maximum performance that can be achieved automatically.

As human assignment is extremely time-consuming, the specialists were only able to provide descriptors for 162 English and for 98 Spanish texts of the test collection. They evaluated a total of 3706 automatically assigned descriptors. These are the basis of the evaluation measures given in **Table 4**.

In the evaluation, the evaluators were given the choice between the choices (a) *good*, (b) *BT* or (c) *NT* (rather good, but a broader or narrower term would have been better), (d) *unknown*, (e) bad, but *semantically related* and (f) *bad*. The results in Table 4 show categories judged with (a) to (c) as correct, all others as incorrect.

The performance is expressed using Precision, Recall and F-measure. For the latter, we gave an equal weight to Recall and Precision. All measurements are calculated separately for each rank. *Precision* for a given rank is defined as the number of descriptors judged as correct divided by the number of descriptors suggested up to this rank. *Recall* is defined as the number of correct descriptors found up to this rank, divided by all descriptors the evaluator found relevant for the text. The evaluator working on English judged an average of 8.15 descriptors as correct (of which 7.5 as (a); standard deviation = 2.5). The person working on the Spanish evaluation accepted a higher average of 11.6 descriptors per text as good (of which 10.4 as (a); standard deviation = 4.2).

## 7.1 Manual evaluation of manual assignment

It is well-known that human indexers do not always come to the same assignment and evaluation results. It is thus obvious that automatically generated results can never be 100% the same as those of a human indexer. In order to have an upper-bound benchmark for our system (i.e. a maximally achievable result), we subjected the previously manually assigned descriptors to the evaluation of our indexing professionals. The review was blind, meaning that the evaluators did not know which descriptors had been assigned automatically and which ones manually.

This manual evaluation of the manual assignment showed that the evaluators working on the English and Spanish texts judged, respectively, 74% and 84% of the previously manually assigned descriptors as good (a). They judged an additional 4% and 3% as rather good ((b) or (c)). Their total agreement with the previously manually assigned descriptors was thus 78% and 87%, respectively. It follows that they actively *disagreed* with the manual assignment in 22% and 13% of all cases. The differences between the two professional and well-trained human evaluators show that there is a difference in style regarding the number of descriptors assigned and regarding the generosity with which they accepted the manually or automatically assigned descriptors as correct.

We take the 74% and 84% overlap with the human judgements for English and Spanish as the maximally achievable benchmark for our system. These inter-annotator agreement results confirm previous studies (e.g. Ferber, 1997 and Jacquemin, 2002), which found an overlap between human indexers of between 20 and 80 percent. 80% can only be achieved by well-trained indexing professionals that are given clear indexing instructions.

## 7.2 Evaluation of automatic assignment

**Table 4** shows human evaluation results for various ranks, i.e. looking at the top-ranking 1, 3, 5, 8, 10 and 11 automatically assigned descriptors. Looking at ranks 8 and 11 is most useful as these are the average numbers of appropriate descriptors as judged by the human evaluators for English and Spanish. Setting the human assignment overlap of 78% and 87% as the benchmark (100%), the system achieved a precision of 86% (67/78) for English and of 80% (69/87) for Spanish. The indexing specialists judged that this is a very good result for a complex application like this one. Regarding the precision values, note that, for those documents for which the professional indexer only found 4 relevant descriptors, the maximally achievable automatic result at rank 8 would be 50% (4/8).

| Nb. of descr | English 162 texts | | | Spanish 98 texts | | |
|---|---|---|---|---|---|---|
| | **P** | **R** | **F** | **P** | **R** | **F** |
| **1** | 94 | 12 | 21 | 88 | 8 | 15 |
| **3** | 83 | 31 | 45 | 86 | 24 | 37 |
| **5** | 75 | 46 | 57 | 82 | 37 | 51 |
| **8** | 67 | 63 | 65 | 75 | 54 | 63 |
| **10** | 58 | 68 | 63 | 71 | 64 | 67 |
| **11** | 55 | 71 | 62 | 69 | 75 | 68 |

**Table 4**. **P**recision, **R**ecall and **F**-measure results for the manual evaluation of the English and Spanish documents of the test collection.

### 7.3 Performance across languages

The system has currently been trained and optimised for English, Spanish and French. According to the automatic comparison with previously manually assigned descriptors, the results were very similar for the three languages. For eight other European languages, assignment was carried out without any linguistic pre-processing and without fine-tuning the stop word lists. The results between languages varied little and were very similar to the results for English, Spanish and French without linguistic input and parameter tuning (which improved the results by six to eight percent). This similar performance across the very different languages (including Finnish and German) shows that the approach as such is language-independent and that the application can easily be applied to further languages when training material becomes available.

### 8 Conclusion and Future Work

The manual evaluation of the automatic assignment of descriptors from the conceptual thesaurus EUROVOC using statistical methods and a high number of optimised parameters showed that the system performed 570% better than the lower bound benchmark, which is the keyword *extrac-*

*tion* of descriptors present verbatim in the text (section 3.2; F-measure comparison). Furthermore, it performs only 14% (English) to 20% (Spanish) less well than the upper bound benchmark, which is the percentage of overlap between two human indexing specialists (section 7.2). Results for English and Spanish assignment were very similar.

We showed that adding a number of parameters to more standard formulae, and identifying the best parameter settings empirically, improves the assignment results a lot. We have successfully applied the language-independent algorithm to more languages, and we believe that it can be applied to other applications such as the indexing of texts with other thesauri. However, the identification of the best parameter setting will have to be done anew. The optimised parameter settings for English, Spanish and French descriptor assignment were similar, but not entirely identical.

A problem we did not manage to solve with different formulae and parameters is the frequent assignment of descriptors that are wrong, but that are clearly part of the same semantic field. For instance, the descriptor NUCLEAR ACCIDENT was often assigned automatically to texts in which vocabulary such as 'plutonium' and 'radioactive' was abundant, even if the texts were not about nuclear *accidents*. Indeed, the descriptors NUCLEAR MATERIAL and NUCLEAR ACCIDENT have a large amount of associates in common, which makes them hard to distinguish. To solve this problem, it is obvious that, for texts on nuclear accidents, the occurrence of at least one of the words 'accident', 'leak', or similar should be made obligatory. We have therefore started applying Machine Learning methods to infer such rules. First experiments with Support Vector Machines are encouraging.

In addition to the assignment in its own right, we use the automatic assignment of EUROVOC descriptors to texts for a variety of other applications. These include cross-lingual document similarity calculation, the automatic identification of document translations, multilingual clustering and classification, as well as subject-specific summarisation.

## Acknowledgements

We would like to thank the Documentation Centres of the European Parliament and of the European Commission's Publications Office OPOCE for providing us with the EUROVOC thesaurus and the training material. We thank Elisabet Lindkvist Michailaki from the Swedish Parliament and Victoria Fernández Mera from the Spanish Senate for their thorough evaluation of the automatic assignment results. We also thank the anonymous evaluators for their feedback given to our initial submission.